\documentclass[runningheads]{llncs}
\usepackage[T1]{fontenc}
\usepackage[utf8]{inputenc}
\usepackage[english]{babel}
\usepackage{amsmath}
\usepackage{amsfonts}
\usepackage{amssymb}
\usepackage{graphicx}
\usepackage{color}
\usepackage{cleveref}
\usepackage{listings}

\DeclareMathOperator*{\argmax}{arg\,max}

\begin{document}

\title{Triadic Concept Analysis for Logic Interpretation of  Simple Artificial Networks}

\author{Ingo Schmitt\orcidID{0000-0002-4375-8677}}
\authorrunning{I. Schmitt}

\institute{Brandenburgische Technische Universität Cottbus-Senftenberg,  Germany \\
\email{schmitt@b-tu.de}\\
\url{https://www.b-tu.de/fg-dbis}
}

\maketitle


\begin{abstract}
An artificial neural network (\texttt{ANN}) is a numerical method used to solve complex classification problems. Due to its high classification power, the \texttt{ANN} method often outperforms other classification methods in terms of accuracy. However, an \texttt{ANN} model lacks interpretability compared to methods that use the symbolic paradigm. Our idea is to derive a symbolic representation from a simple \texttt{ANN} model trained on minterm values of input objects. Based on ReLU nodes, the \texttt{ANN} model is partitioned into cells. We convert the \texttt{ANN} model into a cell-based, three-dimensional bit tensor. The theory of Formal Concept Analysis applied to the tensor yields concepts that are represented as logic trees, expressing interpretable attribute interactions. Their evaluations preserve the classification power of the initial \texttt{ANN} model.

\end{abstract}

\newcommand{\drop}[1]{}
\newcommand{\replace}[2]{#2}

\section{Introduction}
\label{sec:intro}

Besides good accuracy, a classification model should be interpretable. An artificial neural network model, or \texttt{ANN} model, provides good accuracy for many classification tasks but has low interpretability. Numerical methods like an \texttt{ANN} model hide internal decision logic, resulting in a lack of trust \cite{Mil19,KauUslRit22}. Therefore, an \texttt{ANN} is often called a black-box solution.

Interpretability of a classification model is a key requirement in critical applications such as medical diagnosis or decisions that have a strong impact on human life. A task related to interpretation is to explain the classification result for a given input object.

Many researchers, such as those cited in \cite{KraTschWei23,Mol20}, strive to make the black-box behavior of an \texttt{ANN} model more transparent and comprehensible. Most of these methods are not based on logic. For example, the Shapley value \cite{Sha53} and its adaptation SHAP \cite{Lun20} compute the importance of single input attributes as part of an additive measure. The SHAP approach generalizes the local interpretable model-agnostic explanations (LIME) \cite{Rib16}. Another proposal is to extract ridge and shape functions in the context of an extracted generalized additive model to illustrate the effect of attributes. Approaches to measuring the importance of an input attribute can be generalized to measure the strength of interactions among several attributes; see, for example, \cite{MurSon93}.

Note that decision tree classifiers including their derivatives \cite{Sut16} and the Tsetlin machince \cite{Gra19} are based on Boolean logic decisions. Logic expressions, in general,  are seen as much easier to interpret than numerical (sometimes called subsymbolic) classifiers like \texttt{ANN} models. The work from \cite{Bal91,Gar92,Blu04} is an early attempt to bridge neural networks (subsymbolic paradigm) with logic (symbolic paradigm). It defines schemata representing states of a neural network and fuzzy-logic-like operations (conjunction and disjunction). However, these operations do not obey the laws of Boolean algebra. 

The work in \cite{Gra04} discusses the concepts of 'incompatibility' and 'implementation' in the context of bridging the symbolic and subsymbolic paradigms of information processing and relates both paradigms to the concepts of quantum mechanics.

An interesting approach \cite{Bal17,Yan18} to bridge logic in the form of decision trees with a deep neural network is to simulate a decision tree using a neural network. After learning and combining the binnings of attribute values, linear networks are trained for every binning combination in order to learn the classification model.  In contrast, in our work, the starting point is a given trained simple \texttt{ANN} model, which we attempt to interpret \emph{after} training. Furthermore, the partitioning in \cite{Bal17,Yan18} is based on decision tree split nodes simulated by the softmax function, whereas in our approach, it is based on ReLU nodes.

The main idea of our method is to map a trained \texttt{ANN} model to non-Boolean logic trees by leveraging probability theory, Formal Concept Analysis, and the quantum-logic-inspired decision tree (QLDT \cite{Sch22adbis}). We obtain a linear combination of logic trees that helps to interpret the semantics of a given \texttt{ANN} model. This approach resembles the RuleFit method \cite{FriBog08}, which relies on a linear combination of rules derived from decision trees. However, decision trees are based on Boolean logic and lack smoothness with respect to continuous input values.

For our interpretation method,   we define some restrictions attributed to a one-class \texttt{ANN} classifier model already trained using balanced training data $TR$, both of which are the starting point for us:

\begin{itemize}
\item The \texttt{ANN} consists of linear layers (such as convolution, average pooling, and fully connected layers), one layer containing exclusively ReLU nodes, and one output node.
\item The target output of the classification is 1 for the class decision and 0 otherwise. The target value is obtained by applying a threshold operator $\tau$ on the output node.
\item The number $\#atts$ of input object attributes is small. The values of every attribute lie within the unit interval $[0,1]\in \mathbb{R}$, expressing the degree of fulfillment of a human-understandable property. For example, see the left part of Table~\ref{tab:objects} with two training objects $(x_1,y_1), (x_2,y_2)$ and two attributes $a_1, a_2$ with the attribute values $x_i[1],x_i[2]$.
\item The \texttt{ANN} has $2^{\#atts}$ input nodes for the minterm values (defined below) of an input object.
\end{itemize}

\begin{table}
\caption{\label{tab:objects} Two example  objects $(x_1,y_1), (x_2,y_2)$ over two attributes $a_1,a_2$ with $x_i=(x_i[1],x_i[2])$ (left part) and four derived minterms (right part):  $mt_i[0]=(1-x_i[1])\cdot (1-x_i[2]),mt_i[1]=(1-x_i[1])\cdot x_i[2],mt_i[2]=x_i[1]\cdot (1-x_i[2]),mt_i[3]=x_i[1]\cdot x_i[2]$}
\begin{center}
\begin{tabular}{c|cc|c||cccc}
object   &   \multicolumn{2}{c|}{attr.\  values}  & target &\multicolumn{4}{c}{derived minterms}\\
$i$ & $x_i[1]$& $x_i[2]$& $y_i$ & $mt_i[0]$& $mt_i[1]$& $mt_i[2]$& $mt_i[3]$\\\hline
$1$ & 0.8 & 0.1 & 1 & 0.18 & 0.02 & 0.72 & 0.08\\
$2$ & 0.5 & 0.6 & 0 & 0.2 & 0.3 & 0.2 & 0.3
\end{tabular}
\end{center}
\end{table}

\noindent
For a logical interpretation of an \texttt{ANN} model, we require $2^{\#atts}$ minterm values $mt_i[k]$ instead of $\#atts$ values $x_i[j]$ as input. Minterm values are conjunctions (expressed as products) of negated $(1-x_i[j])$ and non-negated attribute values $x_i[j]$.   Each minterm value expresses a specific interaction among all attributes. In accordance with \cite{Sch22adbis,Sch22ideas}, we define:
$$mt_i[k]:=\Pi_{j=1}^{\#atts} \left(1-x_i[j]\right)^{1-b_j}\cdot x_i[j]^{b_j}$$ 
where  the integer $k=0,\ldots,2^{\#atts}-1$ is the minterm identifier and $b=b_1\ldots b_{\#atts}$ is the   bit code of  $k$ ($k=\sum_{j=0}^{\#atts-1}b_{\#atts-j}\cdot 2^j$). It can be shown that $\sum_kmt_i[k]=1$ holds.
For example, see the right part of  Table~\ref{tab:objects} for the $2^2$ minterm values $mt_i[k]$.

\begin{figure}
\centerline{\includegraphics[scale=0.3]{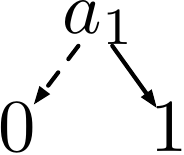}
\hspace{10mm}
\includegraphics[scale=0.3]{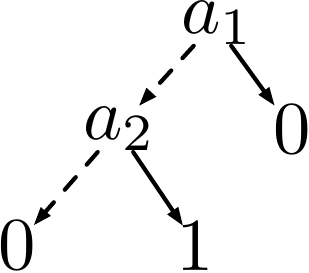}
\hspace{10mm}
\includegraphics[scale=0.3]{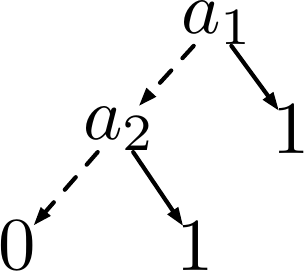}}
\caption{\label{fig:tree} Example logic trees: solid lines refer to non-negated attribute evaluations, while dashed lines refer to negated attribute evaluations. Arithmetic evaluations of the logic trees are as follows: $[a_1]^i=x_i[1]$ (left), $[\overline{a}_1 \land a_2]^i=(1-x_i[1]) \cdot x_i[2]$ (middle), and $[a_1 \lor (\overline{a}_1 \land a_2)]=x_i[1] + (1-x_i[1]) \cdot x_i[2]$ (right).}
\end{figure}

As a result of our method, we obtain logic trees. Three very simple examples, along with their arithmetic evaluations, are depicted in Figure~\ref{fig:tree}.  We will describe such logic trees in more detail later.

\section{ANN Partition Cells}
\label{sec:cells}

A ReLU node in an \texttt{ANN} is defined by the function
 \[ReLU(z)=\left\{\begin{array}{ll}z&\text{if }z\ge 0\\0&otherwise.\end{array}\right.\]
Let us assume the given \texttt{ANN} model has $l$ ReLU nodes, denoted as $ReLU_1, \ldots, ReLU_l$. For every input object, the status of a ReLU node is either \emph{active}, corresponding to the identity function, or \emph{inactive}, corresponding to the zero function. In the following, we will denote the status of an active ReLU node by the bit 1 and an inactive node by the bit 0. Because of the linear layers below a ReLU node, its status separation between active and inactive corresponds to a hyperplane in the input space $[0,1]^{2^{\#atts}}$.

For an input object $x_i$, let us interpret the sequence of status bits of all $l$ ReLU nodes as an integer, which we call its \emph{partition number} $p_i$. We define $active(p)$ as the set of active ReLU nodes for a given partition number $p$. Partition numbers identify \emph{partition cells} of the input space $\mathbb{R}^{2^{\#atts}}$:  input objects with the same partition number belong to the same partition cell. Since every ReLU node corresponds to a hyperplane, every partition cell induces a convex subset of $\mathbb{R}^{2^{\#atts}}$. All partition cells are mutually disjoint. Because all layers of the \texttt{ANN} model except the ReLU nodes are linear, we can combine the linear layers for a given partition cell $p$ into one linear map $mw^{p}$ from the input minterm values to the output node value.  Here, $mw^p[k]$ denotes the minterm weight for minterm $k$. An example of linear mappings of partition cells over $2^2$ minterms is given in Table~\ref{tab:mws}.

\begin{table}
\caption{\label{tab:mws} Example values $mw^p[k]$ for $2^2$ minterms $k=0,\ldots,3$}
\begin{center}
\begin{tabular}{ccccc}
partition   &  \multicolumn{4}{c}{minterm weights}  \\
$p$ & $mw^p[0]$& $mw^p[1]$& $mw^p[2]$& $mw^p[3]$ \\\hline
$1$ & -8 & 3 & 6 & 2\\
$\vdots$ & $\vdots$ & $\vdots$ & $\vdots$ & $\vdots$
\end{tabular}
\end{center}
\end{table}

\begin{figure}
\centerline{\includegraphics[scale=0.3]{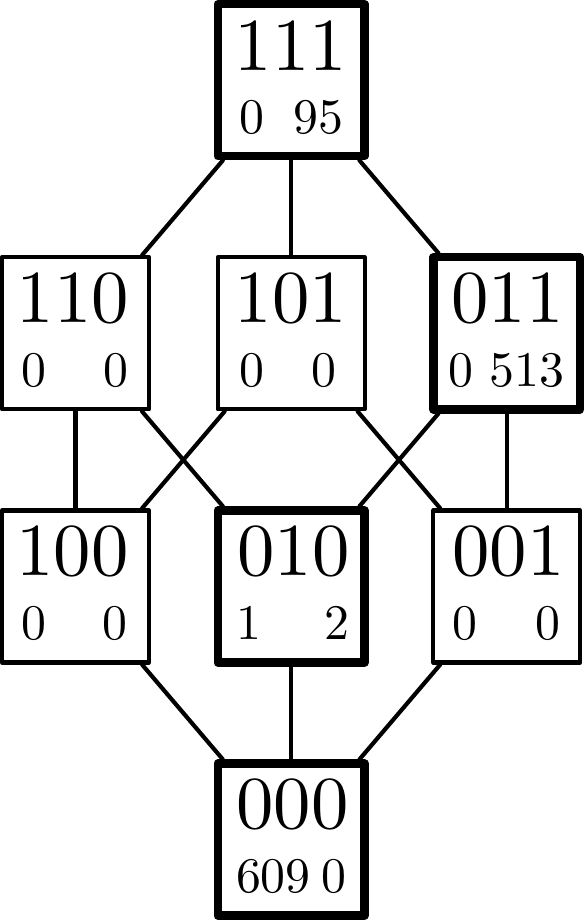}}
\caption{\label{fig:partition-lattice}Example lattice of partition cells identified by bit codes for the status of three ReLU nodes: the small number on the left indicates the number of zero-objects ($y_i=0$), and the small number on the right indicates the number of one-objects ($y_i=1$) in $TR$. Bold partitions are those that are of interest (non-empty).}
\end{figure}

\noindent
For a given \texttt{ANN} model, there are at most $2^l$ partition cells. If a cell $p$ lies completely outside the input space $[0,1]^{2^{\#atts}}$, then there is no input object $x_i$ for which $p_i = p$ holds.  All partition cells form an atomic, Boolean subset lattice over the respective sets $active(p)$.  Figure~\ref{fig:partition-lattice} depicts an example lattice for three ReLU nodes and eight partitions cells.  

For every partition cell we use following information:
\begin{itemize}
\item Number of training objects  $i$ for which $p_i=p$
\item Number of 1-objects and number of 0-objects with $p_i=p$
\item Minterm weigths: $mw^{p}$
\end{itemize}

\noindent
Based on this information, not all partition cells are essential for interpreting an \texttt{ANN} model. Cells with no assigned training objects or cells containing only 1-objects or only 0-objects are of low relevance, as they do not contribute to separating input objects. Cells with many training objects and a balanced occurrence of 1-objects and 0-objects are of high relevance.

As a result of our \texttt{ANN} architecture, all minterm weights of a partition cell can be derived by summing the weights from its respective atomic partition cells. A partition cell is considered atomic if exactly one ReLU node is active:
$$mw^p = \sum_{i:b_i=1}mw^{2^{l-i}}$$
where $b_1 b_2 \ldots b_l$ is the bit code of partition cell number $p$. An example derivation is $mw^{101} := mw^{100} + mw^{001}$. This derivation is very useful because only the minterm values of atomic partition cells — namely, $l \cdot 2^{\#atts}$ values — need to be stored.   In the following sections, we assume $P$ to be the set of essential partition cells.

\section{Converting   Minterm Weights to  a Bit Tensor}
\label{sec:mapping}

Let the minterm weights $mw^p[k]$ for all partition cells $p \in P$ over the minterms $k$ be given, as shown for example in Table~\ref{tab:mws}. The input for the \texttt{ANN} model is an object  $x_i$ with minterm values $mt_i[k]$ for every minterm $k$ and the assigned partition cell $p = p_i$, as shown for example in Table~\ref{tab:objects}. The score $\texttt{ANN}(x_i)$ can be written as the scalar product $\texttt{ANN}(x_i) := \langle mt_i, mw^p \rangle = \sum_k mt_i[k] \cdot mw^p[k]$. For example, using the objects in Table~\ref{tab:objects}
 with $p_1 = p_2 = 1$ and the minterm weights in Table~\ref{tab:mws}, we obtain the scores $\texttt{ANN}(x_1) = 3.1$ and $\texttt{ANN}(x_2) = 1.1$.

If we assume different score values for all objects, their scores can be strictly ordered. For our  interpretation method, we convert the minterm weights $mw^p[k]$ into bit values by applying a strictly monotonic increasing linear function $f$ to them.  Since the scalar product is linear in its arguments, applying $f$ to the minterm weights preserves the strict order of the objects:
\begin{eqnarray*}
\langle mt_i,f(mw^p)\rangle &=&f\left( \langle mt_i,mw^p\rangle\right)\\
\langle mt_{i_1},mw^p\rangle > \langle mt_{i_2},mw^p\rangle &\Leftrightarrow &
f\left( \langle mt_{i_1},mw^p\rangle\right) > f\left( \langle mt_{i_2},mw^p\rangle\right)
\end{eqnarray*}
As consequence,  the application of the modified threshold $f(\tau)$ to the scores $f(\texttt{ANN}(x_i))$ preserves the original class decisions. 

Next, we remove the fractional part from the mapped minterm values $f(mw^p[k])$ and encode the remaining integer as a binary number of $\#bits$ bits. This removal of the fractional part introduces a truncation error. Consequently, if we apply the truncated minterm weights to the input objects, objects with different scores might receive the same scores, which means the strict score ordering becomes a weak order.

\begin{figure}
\centerline{\includegraphics[scale=0.3]{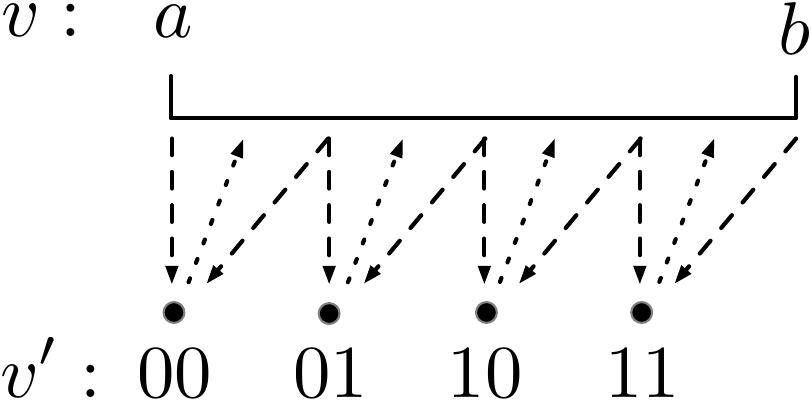}}
\caption{\label{fig:mapping}Dashed lines: lossy mapping of minterm weights to binary numbers \mbox{($\#bits=2$)};}  dotted lines:  reconstruction function
\end{figure}

To determine the linear function $f$, we derive a minterm value interval $[a, b]$ with $a = \min_{p \in P, k} mw^p[k]$ and $b = \max_{p \in P, k} mw^p[k]$. We  encode every minterm weight $v := mw^p[k]$ as an integer $v'$ with $\#bits$ bits:
\begin{eqnarray}
v':=\lfloor f(v)\rfloor :=\left\lfloor \frac{v-a}{b-a}*\frac{2^{\#bits}}{1+\epsilon} \right\rfloor \in \{0,1,\ldots ,2^{\#bits}-1\}.
\end{eqnarray}
The denominator $1 + \epsilon$, where $\epsilon$ is a small value, e.g.  $2^{-(\#bits + 4)}$, is necessary to obtain $2^{\#bits} - 1$ instead of $2^{\#bits}$ for $v := b$. The mapping is depicted in Figure~\ref{fig:mapping}.

Table~\ref{tab:mws1} shows the mapping of our introduced example minterm weights. If we apply the mapped minterm values to the example objects in Table~\ref{tab:objects},  we obtain  the scores $\texttt{ANN}(x_1)'=2.38$ and $\texttt{ANN}(x_2)'=2.1$.

\begin{table}
\caption{\label{tab:mws1} Mapped values $mw^p[k]$  from Table~\ref{tab:mws} over all minterms $k=0,\ldots,3$ of  partition $p=1$ with $\#bits=2$; the last two rows show bit values of $\lfloor f(mv^p[k])\rfloor$  with respect to the two bit levels}
\begin{center}
\begin{tabular}{ccccc}
  &  \multicolumn{4}{c}{Minterms $k$}  \\
& $0$& $1$& $2$& $3$ \\\hline
$mw^p[k]$ & -8 & 3 & 6 & 2\\
$f(mv^p[k])$ & 0 & 3.14 & 3.99 &  2.85\\
$\lfloor f(mv^p[k])\rfloor$ & 0 & 3 & 3 & 2\\
$bl=0:2^0$ & 0 & 1 & 1 & 0\\
$bl=1:2^1$ & 0 & 1 & 1 & 1\\
\end{tabular}
\end{center}
\end{table}

Next, we define a linear reconstruction function that attempts to reconstruct $v$ from $v'$ with a small average reconstruction error. 
The linear reconstruction function for obtaining $v''$ from $v'$, where $v'' \approx v$, is given as:
\begin{eqnarray}
v'':= \left(v'*\frac{1+\epsilon}{2^{\#bits}}+2^{-(\#bits+1)}\right)*(b-a)+a.
\end{eqnarray}
To achieve a small reconstruction error, the term $2^{-(\#bits + 1)}$ selects the middle of the truncation interval, as shown in  Figure~\ref{fig:mapping}.   If we assume a uniform distribution for the minterm values, the average reconstruction error is:
$$E(|v-v''|)=\frac{b-a}{2^{\#bits+2}}$$ and the upper bound is:
$$|v-v''|\le \frac{b-a}{2^{\#bits+1}}.$$ 
Since the reconstruction error decreases exponentially with the number of bits, the parameter $\#bits$ can be small to ensure a small reconstruction error.

After mapping all minterm values, the class decision threshold $\tau$ needs to be mapped as well. The truncation of the minterm values produces an average error of 1/2 on the score values. Thus, we choose $\tau' = f(\tau) - 1/2$.

In our example, we assume the initial value $\tau = 2$ and obtain $\tau' = 2.36$. As a result, the threshold $\tau'$ effectively separates object $x_1$ from $x_2$ after the minterm weight mappings ($\texttt{ANN}(x_1) = 3.1$, $\texttt{ANN}(x_2) = 1.1$, and $\texttt{ANN}(x_1)' = 2.38$, $\texttt{ANN}(x_2)' = 2.1$).

The values $v' = \lfloor f(mw^p[k]) \rfloor$ are regarded as binary numbers. For our example, see the last two rows in Table~\ref{tab:mws1}. Every bit derived from a trained \texttt{ANN} model depends on the combination of a partition cell $p \in P$, a minterm $k \in \{0, 1,\ldots,2^{\#atts} - 1\}$, and a bit level $bl \in \{0, 1,\ldots,\#bits - 1\}$:$$bt(p,k,bl):=\left\lfloor f(mw^p[k])\right\rfloor \&\ 2^{bl}$$
where $bt$ is a three-dimensional bit tensor and $'\&'$  denotes the bitwise \texttt{and}. As a result, the semantics of an \texttt{ANN} model can be approximated by a three-dimensional tensor $bt[p, k, bl]$ of bit values.

From this construction, we can formalize the approximation of $f(\texttt{ANN}(x_i))$ for an input object $x_i$ with $p_i = p$ as:
\begin{eqnarray*}
f(\texttt{ANN}(x_i))=\sum_{k:=0}^{2^{\#atts}-1} mt_i[k] \cdot f(mw^p[k])\approx \sum_{k:=0}^{2^{\#atts}-1} mt_i[k]\cdot \lfloor f(mw^p[k])\rfloor
\end{eqnarray*}
and 
\begin{eqnarray*}
 \sum_{k:=0}^{2^{\#atts}-1} mt_i[k]\cdot \lfloor f(mw^p[k])\rfloor &=&  \sum_{k:=0}^{2^{\#atts}-1} mt_i[k]\cdot\sum_{bl:=0}^{\#bits-1}bt(p,k,bl)\cdot 2^{bl}\\
 &=& \sum_{bl:=0}^{\#bits-1}2^{bl}\cdot\sum_{k:=0}^{2^{\#atts}-1} mt_i[k]\cdot bt(p,k,bl)
\end{eqnarray*}

From a given bit tensor $bt$, we compute its energy:
\begin{eqnarray*}
power(bt)&:=&\sum_{p\in P}\sum_{k:= 0}^{2^{\#atts}-1}\sum_{bl:=0}^{\#bits-1}bt[p,k,bl]\cdot 2^{bl}= \sum_{p\in P}\sum_{k:= 0}^{2^{\#atts}-1}\lfloor f(mw^p[k])\rfloor.
\end{eqnarray*}
The energy can be understood as the contribution of all bit-encoded minterm weights across all partition cells. It approximates the sum of all minterm weights of the given \texttt{ANN} model.

\section{Formal Concept Analysis and Triadic Concept Analysis}
\label{sec:TCA}

Formal Concept Analysis (FCA) is a theory developed by Ganter and Wille \cite{LehWil95,GanWil97,WeiQiaWan18}.  In our method, we use FCA to convert the tensor $bt$ into a compact representation, which we later interpret as logic trees.

Formal Concept Analysis  is based on a binary relation, whereas Triadic Concept Analysis (TCA) can be seen as its generalization to a ternary relation. In the following, we define the main FCA components: A \emph{dyadic context} is a triple $(G,M,I)$ where $G$ is a set of objects and $M$ is a set of attributes\footnote{The terms 'object' and 'attribute' are abstract and do not refer to the input objects and their attribute values in our classification task.}. $I\subseteq G \times M$ is a binary relation over $G$ and $M$. $(g,m)\in I$, also written as $gIm$, indicates that an object $g$ of $G$ has a certain attribute $m$ of $M$. A context can be visualized as a cross table, see for example Table~\ref{tab:context} left, where a cross means $gIm:=(g,m)\in I$. Columns refer to objects, and rows refer to attributes. We define for a subset $A\subseteq G$ the function $':2^G\rightarrow 2^M$ as $A':=\{m\in M|gIm \text{ for all } g\in A\}.$ Analogously, we define for a subset $B\subseteq M$ the function $':2^M\rightarrow 2^G$ as $B':=\{g\in G|gIm \text{ for all } m\in B\}.$ Based on these functions, a \emph{concept} of a context is a pair $(A,B)$ with $A\subseteq G, B\subseteq M,  A'=B,$ and $B'=A$. In other words, $A\times B\subseteq I$ (cross-product rule) holds, and $A$ as well as $B$ cannot be extended without violating the cross-product rule. If we depict a context as a cross table, then a concept $(A,B)$ corresponds to a maximal rectangle filled with crosses for some permutation of columns and rows, see for example the first concept in Table~\ref{tab:context}.
\begin{table}
\begin{center}
\caption{\label{tab:context} Example context and some concepts;  bold crosses correspond to the first concept}
\begin{tabular}{c||c|c|c}
 & $g_1$ & $g_2$ & $g_3$\\\hline\hline
 $m_1$ &  & $\times$ &\\\hline
 $m_2$ && $\boldsymbol{\times}$ & $\boldsymbol{\times}$ \\\hline
 $m_3$ & $\times$ & $\boldsymbol{\times}$ & $\boldsymbol{\times}$ \\\hline
\end{tabular}
\hspace{1cm}
$concepts=\left\{\begin{array}{c}
\boldsymbol{(\{g_2,g_3\},\{m_2,m_3\})}\\(\{g_1,g_2,g_3\},\{m_3\})\\(\{g_2\},\{m_1,m_2,m_3\})
\end{array}\right\}$
\end{center}
\end{table}

For a given context, up to $2^{\min\{|G|,|M|\}}$ concepts can exist. All concepts together form a Boolean subset lattice over the partially ordered set $(\{(A'',A')|A\subseteq G\},\leq)$ with $(A''_1,A'_1)\leq (A''_2,A'_2) \Leftrightarrow A''_1\subseteq A''_2$. Concepts can overlap, e.g., $(\{g_2\},M)$ and $(G,\{m_3\})$ overlap with $g_2Im_3$, meaning they can share crosses.

We derive a set $\texttt{concepts}$ (which does not necessarily contain all concepts) from singleton sets as follows: 
$$\texttt{concepts}:=\{(\{a\}'',\{a\}')|a\in G\}\cup \{(\{b\}',\{b\}'')|b\in M\}.$$
This results in no more than $|G|+|M|$ concepts. Note that every cross $(g,m)\in I$ is covered by at least one concept from $\texttt{concepts}$.
 
Let us now generalize Formal Concept Analysis to Triadic Concept Analysis (TCA):
A \emph{triadic context} is a quadruple $(K_1, K_2, K_3, Y)$ where $K_1$ is a set of objects, $K_2$ is a set of attributes, and $K_3$ is a set of conditions. $Y \subseteq K_1 \times K_2 \times K_3$ is a ternary relation over $K_1$, $K_2$, and $K_3$. The expression $(g, m, b) \in Y$ means that object $g$ has attribute $m$ under condition $b$.
We define for a triadic context, $X_2 \subseteq K_2$, and $X_3 \subseteq K_3$, the function $^1: 2^{K_2} \times 2^{K_3} \rightarrow 2^{K_1}$:
$(X_2,X_3)^1:=\{x_1\in X_1|(x_1,x_2,x_3)\in Y\text{ for all } (x_2,x_3)\in X_2\times X_3\}.$
Analogously, we define the functions $(X_1, X_3)^2$ and $(X_1, X_2)^3$ for computing $X_2$ and $X_3$, respectively. A \emph{triadic concept} is a triple $(X_1, X_2, X_3)$ for which
$X_1 = (X_2, X_3)^1$, $X_2 = (X_1, X_3)^2$, and $X_3 = (X_1, X_2)^3$ hold.
 A triadic concept can be seen as a maximal cuboid fulfilling the cross-product rule $X_1 \times X_2 \times X_3 \subseteq Y$. From the definition, we see that a triadic concept is uniquely defined for two given sets $X_i, X_j$. From a given triadic context $(K_1, K_2, K_3, Y)$, we compute triadic concepts \texttt{triconcepts} slice-wisely by:
$$\texttt{triconcepts}:=\bigcup_{x_1\in K_1} \{(X_2,X_3)^1,X_2,X_3)|(X_2,X_3)\in\texttt{concepts}^{x_1}\},$$
where $\texttt{concepts}^{x_1}$ are the dyadic concepts $\texttt{concepts}$ computed from the $x_1$-slice dyadic context $(G, M, I)$ derived from the triadic context: $(x_2, x_3) \in I \Leftrightarrow (x_1, x_2, x_3) \in Y$ and $G := K_2, M := K_3.$

\section{Generating Logic Trees from Exclusive Triconcepts}
\label{sec:trees}

In  Section~\ref{sec:mapping}, we derived a three-dimensional tensor $bt$ from an \texttt{ANN} model.   We interpret it as a triadic context $(K_1, K_2, K_3, Y)$ with $K_1 = P$, $K_2 = \{0, 1, \ldots, 2^{\#atts}-1\}$, $K_3 = \{0, 1, \ldots, \#bits-1\}$, and $Y = \{(p, k, bl) | bt[p, k, bl] = 1\}$. For example, we can regard the last two rows of Table~\ref{tab:mws1} of the partition $p$ as a triadic context: $K_1 = \{p\}, K_2 = \{0, 1, 2, 3\}, K_3 = \{0, 1\}$, and $$Y=\{(p,1,0),(p,2,0),(p,1,1),(p,2,1),(p,3,1)\}.$$ We compute triadic concepts as $\{c_j(X_1, X_2, X_3)\} := \texttt{triconcepts}$. From our example in Table~\ref{tab:mws1}, we obtain two concepts: $c_1 := c_1(\{p\},\{1, 2, 3\}, \{1\})$ and $c_2 := c_2(\{p\}, \{1, 2\}, \{0, 1\})$. A concept can be seen as a maximal cuboid completely filled with set bits. Analogous to the energy of a context, we now define the \emph{energy} of a concept  $c_j$ as the product of the number of its partition cells, the number of its minterms, and the $power   sum:=\sum_{x_3 \in X_3} 2^{x_3}$ over its bit code levels:
\begin{eqnarray*}
power(c_j(X_1,X_2,X_3))&:=&|X_1|*|X_2|*\sum_{x_3\in X_3}2^{x_3}
\end{eqnarray*}
The energy of our two example concepts is: $power(c_1)=power(c_2)=6.$
In combination with the context energy, we define the \emph{relative concept energy} as $relpower(c):=power(c)/power(bt)\footnote{Alternatively, we can use the number of $TR$ objects (support) of the covered partition cells instead of the number $|X_1|$ of cells for both energy formulas.}.$ 

We will interpret the minterms $X_2$ of a triadic concept as a logic expression with a relative power value computed from $X_3$ and valid for objects of the partition cells in $X_1$.  As mentioned in the previous section, the cuboids of different triadic concepts can overlap and would produce overlapping logic expressions.
Our aim is to obtain non-overlapping logic expressions with high energy. Therefore, we compute an array $ex\_c$ of exclusive triadic concepts ordered by descending relative energy from a triadic context. Thus, $\sum_{xc\in ex\_c}relpower(xc)=1$
holds, or in other words, every set bit of $bt$ is contained in exactly one exclusive triadic concept $xc$.   See Algorithm~\ref{alg:triconcepts} for computing the required array $ex\_c$.  From our example in Table~\ref{tab:mws1} we obtain: $ex\_c[0]=c_0(\{p\},\{1,2\},\{0,1\})$ with $power(ex\_c[0])=6$ and $ex\_c[1]=c_1(\{p\},\{3\},\{1\})$ with $power(ex\_c[1])=2$.

\begin{figure}[h]
\begin{lstlisting}[frame=single,mathescape=true,numbers=left,
    stepnumber=1]
function excl_triconcepts(tricontext bt): array of triconcepts
  ex_triconcepts := []    // $\text{\em empty result array}$
  C := $\texttt{triconcepts}$(bt)
  xc:=$\argmax_{c\in C}$relpower(c)         // $\text{\em concept with highest power}$
  ex_triconcepts.append(xc)
  // $\text{\em reduce tricontext by concept with highest power}$
  for $x_1\in $xc.$X_1$ 
    for $x_2\in $xc.$X_2$
      for $x_3\in $xc.$X_3$
        bt[$x_1,x_2,x_3$] := 0  // $\text{\em no cross}$
  if power(bt) > 0  // $\text{\em is tricontext non-empty}$
    go to line 3
  return ex_triconcepts
\end{lstlisting}
\caption{Extracting exclusive triconcepts}
\label{alg:triconcepts}
\end{figure}

In the algorithm, we used $relpower$ in Line 4 to find the best triadic concept. However, this is only the first of four possible methods:
\begin{enumerate}
\item M1: relative power ($\#partitions * \#minterms * power sum$)
\item M2: $\#partitions * power sum$
\item M3: $power sum$
\item M4: accuracy across the objects of the covered partition cells 
\end{enumerate}
We will compare their impacts on the logic tree generation  in Section~\ref{sec:experiments}.

A set $X_2$ of minterm identifiers from a triadic concept $c_j(X_1,X_2,X_3)$ is not an intuitive representation of a logic expression.  Therefore, we construct a quantum-logic-inspired decision tree (QLDT \cite{Sch22adbis}) from $X_2$ and present it to the expert for interpretation.   The QLDT is a compact representation of a logic expression defined by $X_2$.  The main construction idea is to use the bit codes of active and inactive minterms 
\begin{eqnarray*}
\{(\text{bit-code}(k),1)|k\in X_2\}\cup  \{(\text{bit-code}(k),0)|k\in K_2\setminus X_2\}
\end{eqnarray*} 
as training data for learning a classical decision tree. The bits of the bit code for minterm $k$ represent all $n$ original attributes,  either negated or non-negated.  The attributes near the root of the resulting decision tree are more effective at separating active from inactive minterms than those further away.

\begin{figure}
\centerline{\includegraphics[scale=0.3]{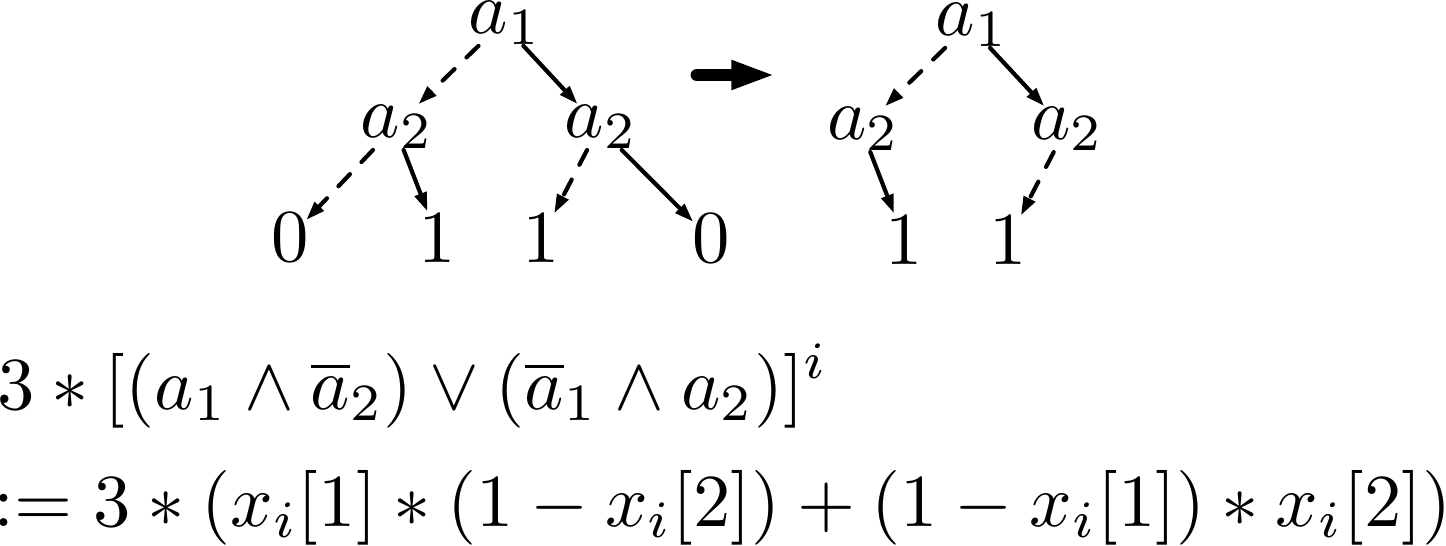}}
\caption{\label{fig:tree1} Example logic tree for $ex\_c[0]$; solid lines non-negated and dashed line means negated evaluation, paths to 0-leaves can be dropped; below the arithmetic evaluation  for an object $x_i$ (discussed in Section~\ref{sec:interpretation})}
\end{figure}

From the example triadic concept
 $ex\_c[0]$ with $X_2=\{1,2\}$ we obtain training data: $\{(00,0),(01,1),(10,1),(11,0)\}$.  The $power sum$ derived  from $X_3$  is $2^0+2^1=3$.  The final logic tree  is depicted in Figure~\ref{fig:tree1}.
 
\begin{figure}
\centerline{\includegraphics[scale=0.3]{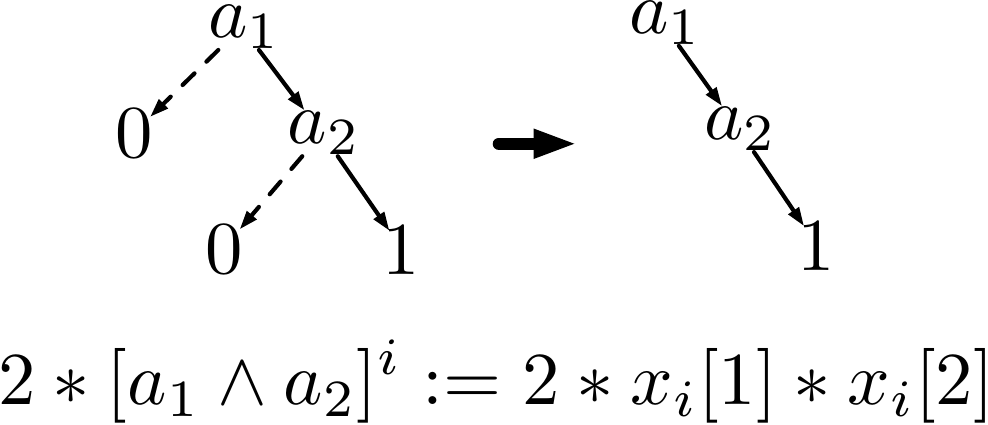}}
\caption{\label{fig:tree2} Example logic tree for $ex\_c[1]$; solid lines non-negated and dashed line means negated evaluation, paths to 0-leaves can be dropped; below the arithmetic evaluation for an object $x_i$}
\end{figure}

Figure~\ref{fig:tree2} depicts the logic tree from   $ex\_c[1]$ with $X_2=\{3\}$, training data  $\{(00,0),(01,0),(10,0),(11,1)\}$  and $power sum$  $2^1=2$.

\section{Evaluation and Interpretation of Logic Trees}
\label{sec:interpretation}

The minterms of a triadic concept can be seen as minterms of a logic expression in disjunction normal form. Since the minterm values of the input objects are continuous, the evaluation of a logic expression following our method is not based on Boolean values. However, as proven in \cite{schmitt2008qql,Sch19}, the underlying commuting quantum logic obeys the rules of Boolean algebra and probability theory on statistically independent attribute values. The evaluation result is a continuous score value that is mapped to a class decision using a threshold operator (a step function) $\tau$.

A logic tree is a compact representation of a logic expression based on minterms that provides good interpretability. An inner tree node branches on a single attribute, where every edge to a child corresponds either to the negated or the non-negated attribute evaluation. Similar to decision trees, leaves refer to class decisions. For our one-class classification problem, we ignore 0-leaves and focus on 1-leaves only.\footnote{Focusing on 0-leaves is the logical negation of focussing on 1-leaves.} Let $L_1$ be the set of all paths from the root to a 1-leaf. All literals on a given path $l\in L_1$, negated or non-negated, are combined conjunctively. Every attribute of an $L_1$-path can occur only once. The branching of an inner node based on an attribute to two children, one negated and the other non-negated, corresponds to an exclusive disjunction.

Following the rules of probability theory, we evaluate a logic tree using three rules:
\begin{enumerate}
\item The probability value of a negated event is computed by subtracting it from 1.
\item The probability value of a conjunction of independent events is computed by the arithmetic product.
\item The probability value of an exclusive disjunction is computed by the arithmetic sum.
\end{enumerate}
As a result, we obtain the evaluation of a logic tree $qldt$ against an object $x_i$ as $[qldt]^i:=\sum_{l\in L_1}\prod_{e_{lj}\in path(l)}[e_{lj}]^i,$ where $[e_{lj}]^i:=x_i[j]$ holds for a non-negated attribute $a_j$ and $[e_{lj}]^i:=1-x_i[j]$ holds for a negated one.
Let us assume a special input object where only  values 0 and 1 occur as attribute values. In that case, the $qldt$ evaluation coincides with Boolean logic. Thus, our evaluation formula  $[qldt]^i$ can be seen as a generalization of Boolean logic.

From several exclusive concepts, we obtain several logic trees. To evaluate an input object $x_i$, we choose the logic trees for concepts that cover the object:
$$QLDT(i)=\{qldt(c)| c\in \texttt{excl\_triconcepts(bt)}\land p_i\in c.X_1\},$$ 
sum their weighted tree evaluations up and obtain the score value:
$$score(x_i)=\sum_{qldt(c)\in QLDT(i)}powersum(c)*[qldt(c)]^i.$$
If $score(x_i)>\tau',$ then we obtain the class value 1; otherwise, we obtain 0.

Due to our mapping from the bit tensor to the logic trees, we obtain for input object $x_i$ with $p=p_i$ the following equivalence:
$$\sum_{qldt(c)\in QLDT}powersum(c)*[qldt(c)]^i=
\sum_{bl:=0}^{\#bits-1}2^{bl}\cdot\sum_{k:=0}^{2^{\#atts}-1} mt_i[k]\cdot bt(p,k,bl).$$
For our example objects, the weighted sum over $ex\_c[0]$ and $ex\_c[1]$ yields 
the $score(x_1)=2.38$  and the $score(x_2)=2.1$. That is, the weighted sum over the logic trees approximates the minterm evaluation.

\subsection*{Interpretation of Minterm Weights and Logic Trees} 

For interpretation,  an expert of the domain and an expert of logic and classification should work together. We will call them below as \emph{expert}.

Based on the minterm weights of a partition cell, we can compute the contribution of individual attributes across the minterms using the Shapley formula \cite{Rot88}:
$$Sh_i=\sum_{S\subseteq N\setminus \{i\}}\frac{(n-1-|S|)!|S|!}{n!}\left(v(S\cup \{i\})-v(S)\right).$$
The main idea is to interpret all non-negated attributes of a minterm as a set. The input for computing the Shapley values are the set of indices $N:=\{1,\ldots,n\}$ for the object attributes $a_1,\ldots,a_n$ and the minterm values $v(S):=mw^p[k]$, where \mbox{$k=\sum_{i\in S}2^{i-1}$} is composed of attributes that are regarded as non-negated attributes of minterm $k$. Please note that $v$ in our case is not necessarily monotonic, and $v(\emptyset)=mw^p[0]$ does not need to be zero. By summing up all Shapley values, we obtain:$$\sum_iSh_i=v(N)-v(\emptyset)=mw^p[2^n-1]-mw^p[0].$$
The Shapley values computed as described above are independent of objects. If we want them to be dependent on an input object $x_i$, then compute $v(S):=mt_i[k]\cdot mw^p[k].$ A general disadvantage of Shapley values is the lack of showing interactions between attributes.

Our interpretation method derives logic trees from an \texttt{ANN} model and is therefore a model dependent and inherent interpretation method.  A logic tree represent complex interactions between attributes for solving a classification problem.

The logic expression behind the tree can be compared with another logic expression that is a hypothesis of the expert. For comparison, the sets of underlying minterms can be analysed on intersections . For example, if every hypothesis minterm is contained in the tree expression, then the hypothesis implies the tree expression. For a deeper discussion, see \cite{Sch22adbis}.

If a derived logic tree $qldt(c)$ is too complex for human interpretation, then its individual  paths to the leaves $l\in L_1$ can be taken for interpretation.  We know from the tree evaluation that the individual path evaluations are simply summed up.  Each individual leaf path can be interpreted as a conjunction of literals.  The following properties of a leaf path $l\in L_1$,  as part of the tree, are independent of TR and are valuable for interpretation:
\begin{itemize}
\item Leaf depth: the  depth $depth(l)$ within the tree   corresponds to the number of path attributes and tells us how many minterm are covered: $2^{n-depth(l)}.$ Here, $n$ is the total number of attributes.
\item Weight of the tree: $powersum(c)$
\item Number of partition cells: $|c.X_1|$
\end{itemize}
For a given input object $x_i$ with $p=p_i$, the path evaluation shows its contribution $[l]^i:=\prod_{j\in path(l)}[a_{lj}]^i$  to the score.  Finally, if we take training (or test) data $T$,  we can compute the average path contribution  $\frac{1}{|T|}\sum_{(x_i,target)\in T} [l]^i$ for the  value $target=0$ or $target=1$ to the scores,   as well as values for accuracy,  recall, and precision.  Examples in an experimental context are given next.

\section{Experimental Interpretation of an \texttt{ANN} Model}
\label{sec:experiments}

The experimental dataset is the blood transfusion service center dataset\footnote{\url{https://archive.ics.uci.edu/ml/datasets.php}}. A classifier needs to be developed to predict whether a person donates blood or not. To make that decision for each person, we have the following information: recency r (months since the last donation), frequency f (total number of donations), monetary status m (total blood donated in c.c.), and time t (months since the first donation). In the balanced case, 178 people donated blood and 178 did not. 

\begin{figure}
\centerline{\includegraphics[scale=0.2]{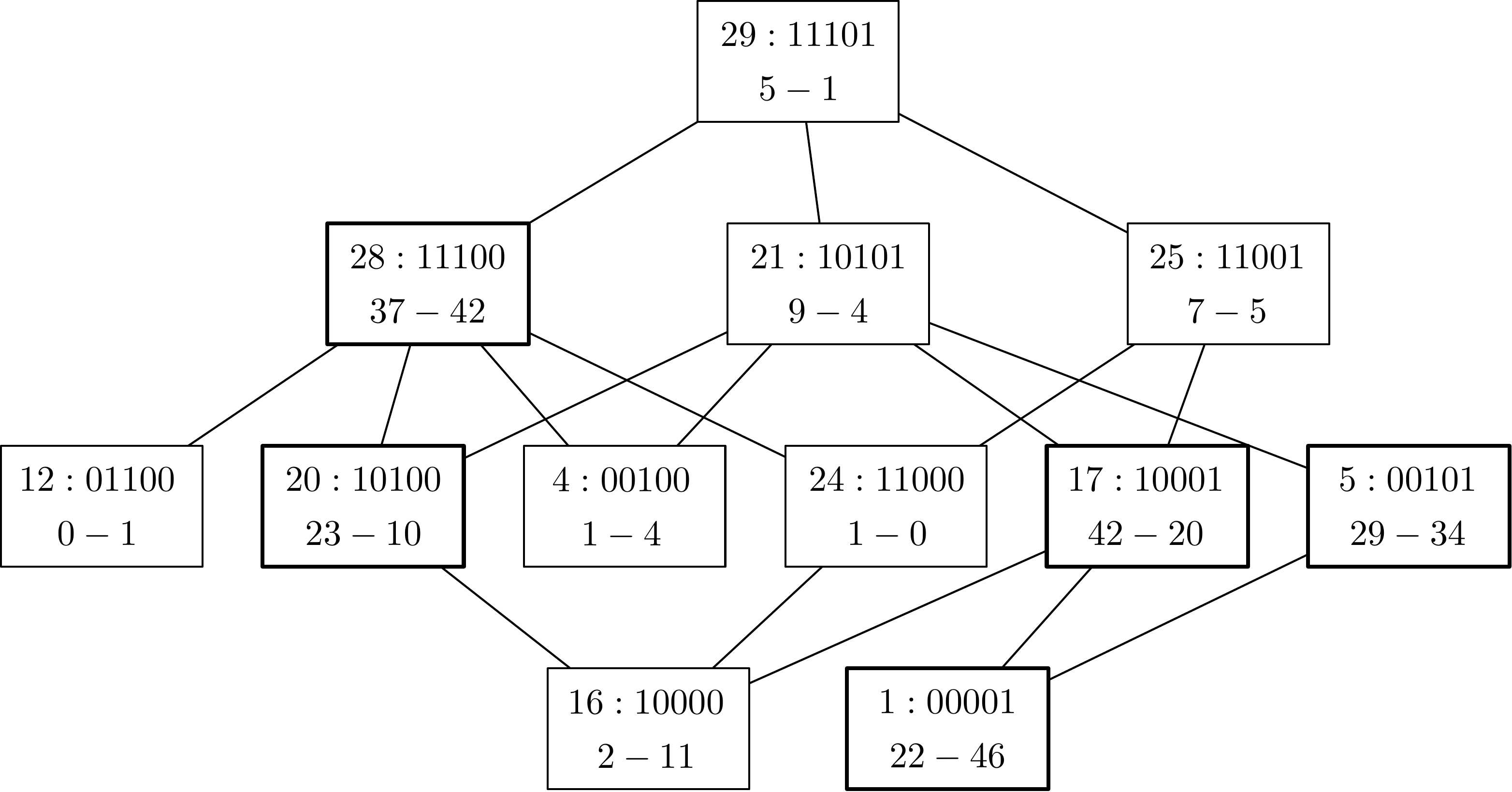}}
\caption{\label{fig:trans-pverband} Transfusion: partially ordered set of non-empty partition cells; bit code shows activity of the ReLU-nodes;  subset relation between sets of active ReLU-nodes form a partially ordered set; lower left value shows the number of 1-objects; lower right number shows the number of 0-objects; selected cells (1,5,17,20,28) have high support (sum of lower numbers) and are indicated as bold-lined boxes}
\end{figure}

We trained an \texttt{ANN} model with 5 ReLU nodes, 16 input nodes, and one output node. The final decision threshold $\tau$ is determined by choosing the training score value that achieves the best accuracy. The transfusion problem is a difficult classification problem, and our model yields 74\% accuracy. From the network, we derived its partition cells. Figure~\ref{fig:trans-pverband} shows a Hasse diagram of the partially ordered set of its non-empty partition cells. We selected partition cells (1, 5, 17, 20, 28) as essential cells: 305 out of the total 356 people belong to the essential cells.

\begin{table}
\caption{Shapley values of the four attributes computed from  the minterms of essential partition cells; the largest absolute values are marked in bold text \label{tab:shapley}}
\begin{center}
\begin{tabular}{c|rrrr}
partition cell & recency & frequency & monetary & \mbox{\hspace{5mm}}time\\\hline
1 & \textbf{0.044} & $-0.026$& $-0.026$& $0.008$\\
5 &\textbf{-0.075}& 0.041& $0.032$& $0.021$\\
17 &\textbf{0.062}& $0.001$& $0.008$& $0.049$\\
20 & \textbf{-0.101}& $0.095$& $0.093$& $0.054$\\
28 & $0.081$& \textbf{-0.119}& $0.011$& $0.005$
\end{tabular}
\end{center}
\end{table}

Table~\ref{tab:shapley} shows the Shapley values of the essential partition cells. The attribute \texttt{time} is positive across all essential partition cells but shows low impact. The attributes \texttt{recency} and \texttt{frequency} have a high impact; however, they are positive in some partition cells and negative in others. This indicates that these attributes play different roles depending on other attributes (high interactions), and the behavior of people in different cells is not homogeneous.

\begin{table}
\caption{Top exclusive concepts using methods M1, M2, M3,   and M4\label{tab:concepts}}
\begin{center}
\begin{tabular}{cccc}
concept & partition cells & minterms & bit levels  \\
\hline
\multicolumn{4}{l}{M1: $\#partitions * \#minterms * power sum$}\\\hline
$c_1$&$\{1,5,17\}$ &$\{0,2-7,9,15\}$ &$\{1\}$ \\
$c_2$&$\{28\}$ &$\{0-2,4,8,9,11,13\}$ &$\{0\}$ \\
$c_3$&$\{20,28\}$ &$\{1,6-8,14,15\}$ &$\{1\}$ \\
\hline
\multicolumn{4}{l}{M2: $\#partitions  * power sum$}\\\hline
$c_4$&$\{5,20,28\}$ &$\{8\}$ &$\{0,4\}$ \\
$c_5$&$\{5,17,20\}$ &$\{14\}$ &$\{0\}$ \\
$c_6$&$\{1,17,28\}$ &$\{1\}$ &$\{0\}$  \\
\hline
\multicolumn{4}{l}{M3: $power sum$}\\\hline
$c_7$&$\{5,20,28\}$ &$\{8\}$ &$\{0,4\}$ \\
$c_8$&$\{17\}$ &$\{1,14\}$ &$\{0,6\}$ \\
$c_9$&$\{5,20\}$ &$\{14\}$ &$\{0,3\}$  \\
\hline
\multicolumn{4}{l}{M4: accuracy}\\\hline
$c_{10}$&$\{20\}$ &$\{2, 4, 5, 9-14\}$ &$\{5\}$ \\
$c_{11}$&$\{17\}$ &$\{1, 2, 4, 6-13, 15\}$ &$\{2\}$ \\
$c_{12}$&$\{5\}$&$\{2, 4-6, 10-14\}$&$\{3\}$\\
$c_{13}$&$\{1\}$&$\{0, 2-9, 14, 15\}$&$\{1\}$\\
$c_{14}$&$\{1,20,28\}$&$\{6\}$&$\{3\}$
\end{tabular}
\end{center}
\end{table}

\begin{figure}
\centerline{\includegraphics[scale=0.3]{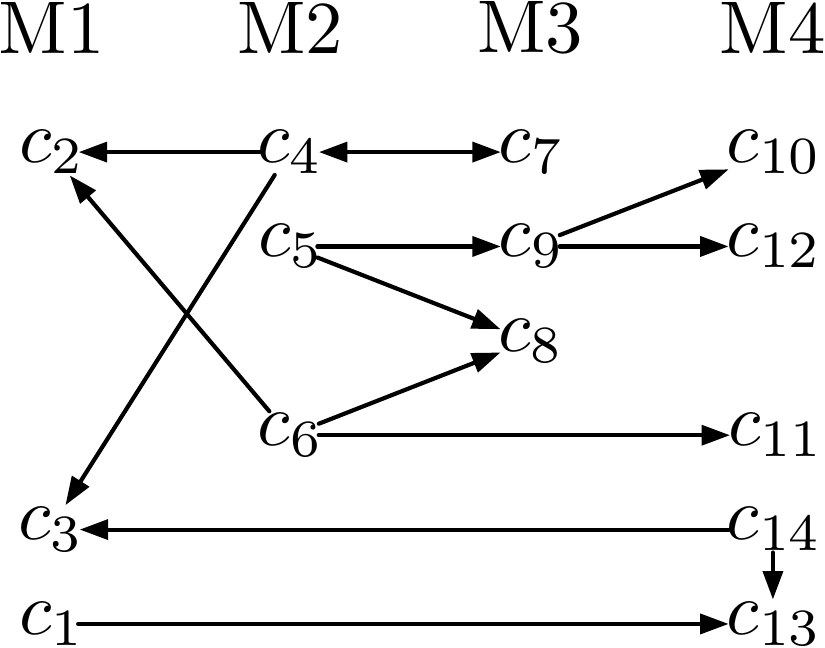}}
\caption{\label{fig:concept-poset} Transfusion: concepts and their implications
(subsets of minterm sets and supersets of partition cell sets)}
\end{figure}

From the minterm values, we compute the bit tensor with 7 bits. Testing the resulting concepts yields an accuracy of 74\%, indicating that the truncation error is small enough. Table~\ref{tab:concepts} shows the top concepts based on different methods of selecting the best concepts. Comparing the concepts derived from different methods reveals implications among them, as illustrated in Figure~\ref{fig:concept-poset}.  One concept implies another if their minterm sets are in a subset relation and their partition cell sets are in a superset relation. Although concepts $c_{13}$ and $c_{14}$ are exclusive due to their distinct sets of bit levels, concept $c_{14}$ implies concept $c_{13}$ because of the subset relation between their minterm sets.

\begin{table}
\caption{Concept characteristics\label{tab:concepts_characteristics}}
\begin{center}
\begin{tabular}{ccccccc}
concept &  $relpower$ & precision & recall & accuracy & support \\\hline
$c_1$ &\textbf{19\%} & 58\% & 26\% & 55\% & \textbf{193}\\
$c_2$  &11\% & 51\% & 54\% & 54\% & 79\\
$c_3$ &8\% & 64\% & 26\% & 52\% & 112\\\hline
$c_{4}$ & 4\% &  53\% &  98\% &  54\% & 175\\
$c_{5}$ & 4\% &  75\% &  50\% &  60\% & 158\\
$c_{6}$ & 4\% &  \textbf{100\%} &  1\% &  52\% & 209\\\hline
$c_{7}$ & 4\% &  53\% &  98\% &  54\% & 175\\
$c_{8}$ & 2\% &  75\% &  73\% &  66\% & 62\\
$c_{9}$ & 3\% &  63\% &  84\% &  65\% & 96\\\hline
$c_{10}$ & 0.4\% &  71\% &  \textbf{100\%} & \textbf{72\%} & 33\\
$c_{11}$ & 4\% &  77\% &  83\% &  \textbf{72\%} & 62\\
$c_{12}$ & 1\% & 76\% &  55\% &  71\% & 63\\
$c_{13}$ & 1\% & 70\% &  97\% &  70\% & 68\\
$c_{14}$ & 7\% & \textbf{100\%} &  9\% &  70\% & 180\\
\end{tabular}
\end{center}
\end{table}

Next, we examine the properties of the derived top concepts in Table~\ref{tab:concepts_characteristics} with regard to $relpower$, precision, recall, accuracy, and support (number of covered training objects). There are very specific concepts ($c_6,c_{14}$) with high precision and low recall, but also very general concepts ($c_4,c_7,c_{10},c_{13}$) with high recall and low precision. Due to the method of selecting the best concepts, method M1 concentrates concepts with high $relpower$ at the top, whereas method M4 prefers concepts with high accuracy. Ultimately, the expert must decide which method is preferred.

Focusing on M4-concepts, we see that the concepts $c_{10},c_{11},c_{12},c_{14}$ cover all essential partition cells. At the same time, their accuracy values are not less than 70\%, which is close to the network's accuracy of 74\%. The support is 305 out of the total 356. This means that only four concepts are necessary to approximate the accuracy of the network.

\begin{figure}
\centerline{\includegraphics[scale=0.3]{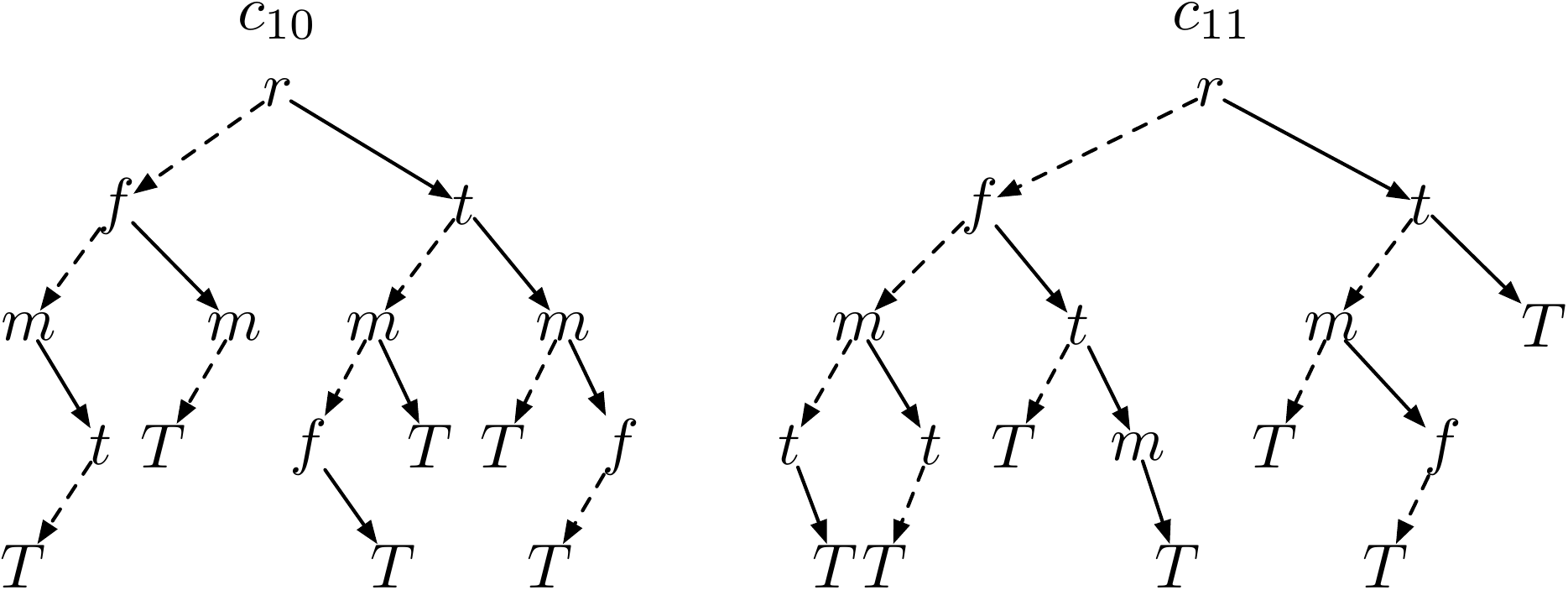}}
\caption{\label{fig:trans-trees} Transfusion: logic trees for concepts $c_{10}$ and $c_{11}$  over essential cells,  $T$ stands for transfusion class,  and a dashed line indicates a negated evaluation of the previous attribute}
\end{figure}

\begin{table}
\caption{Leaf paths of the first concept logic trees and their characteristics; bold numbers indicate high values\label{tab:leaves}}
\begin{center}
\begin{tabular}{lcccccc}
leaf path & precision & recall & accuracy & \#mt & 0-avg & 1-avg \\\hline
$c_{10}$\\\hline
$\overline{r}\land \overline{f}\land m\land\overline{t}$ &75\%&65\%&60\%& 1 &0.049&0.047\\
$\overline{r}\land {f}\land \overline{m}$ &\textbf{100\%}&13\%&39\%&2&0.125&0.112\\
${r}\land {f}\land \overline{m}\land\overline{t}$ &71\%&\textbf{100\%}&\textbf{72\%}&1&0.023&0.025\\
${r}\land m\land\overline{t}$ &83\%&43\%&54\%&2&0.080&0.085\\
${r}\land \overline{m}\land {t}$ &\textbf{100\%}&13\%&39\%&2&0.056&0.047\\
${r}\land \overline{f}\land m\land {t}$ &71\%&\textbf{100\%}&\textbf{72\%}&1&0.040&0.030\\\hline
$c_{11}$ \\\hline
$\overline{r}\land \overline{f}\land \overline{m}\land {t}$ &73\%&71\%&62\%&1&0.025&0.023\\
$\overline{r}\land \overline{f}\land m\land\overline{t}$ &72\%&64\%&59\%&1&0.027&0.020\\
$\overline{r}\land {f}\land\overline{t}$ &67\%&\textbf{100\%}&67\%&2&0.074&0.050\\
$\overline{r}\land {f}\land m\land {t}$ &\textbf{100\%}&11\%&40\%&1&0.271&0.193\\
${r} \land \overline{m}\land\overline{t}$ &75\%&73\%&66\%&2&0.065&0.131\\
${r}\land \overline{f}\land m\land\overline{t}$ &76\%&69\%&64\%&1&0.034&0.054\\
${r}\land {t}$ &75\%&71\%&64\%&4&0.336&0.382\\
\end{tabular}
\end{center}
\end{table}

Let us now examine the logic interpretations of the concepts. The concept $c_{14}$, with just one minterm, is very simple: $\overline{r} \land f \land m \land \overline{t}$. Figure~\ref{fig:trans-trees} depicts the logic trees of the concepts $c_{10}$ and $c_{11}$. Since these are hard to interpret, we examine their paths to 1-leaves, as shown in Table~\ref{tab:leaves}. For each path, we present precision, recall, accuracy, the number of covered minterms (\#mt), and the averaged path evaluations, separated for 0-objects (0-avg) and 1-objects (1-avg). The difference between 0-avg and 1-avg can be seen as a measure of the separation power. Notably, the third and sixth leaf paths of $c_{10}$ are interesting because their accuracy of 72\% is very effective for classification.

Table~\ref{tab:leaf-implications} shows the implications between the leaf paths listed in Table~\ref{tab:leaves}. These implications together with Table~\ref{tab:leaves}  illustrate how adding minterms to a concept affects its characteristics.

\begin{table}
\caption{Implications between  leaf paths  of concepts $c_{10}$ and $c_{11}$\label{tab:leaf-implications}}
\begin{center}
\begin{tabular}{lcl}
$c_{10}$ leaf   & implication & $c_{11}$ leaf \\\hline
$\overline{r}\land \overline{f}\land m\land\overline{t}$ & $\longleftrightarrow$ & $\overline{r}\land \overline{f}\land m\land\overline{t}$\\
${r}\land {f}\land \overline{m}\land\overline{t}$ & $\longrightarrow$ & ${r} \land \overline{m}\land\overline{t}$\\
${r}\land m\land\overline{t}$ &$\longleftarrow$ & ${r}\land \overline{f}\land m\land\overline{t}$\\
${r}\land \overline{m}\land {t}$&$\longrightarrow$ &${r}\land {t}$ \\
${r}\land \overline{f}\land m\land {t}$&$\longrightarrow$ &${r}\land {t}$ 
\end{tabular}
\end{center}
\end{table}

\section{Conclusion}
\label{sec:conclusion}

We addressed the problem of interpreting  \texttt{ANN} models using logic expressions by constructing a bridge between them and visualizing these expressions as logic trees (QLDT). Thus, our interpretation method decomposes an \texttt{ANN} model into logic trees.

ReLU nodes in an \texttt{ANN} introduce non-linearity to the network, which is essential for solving non-linear classification tasks. Our approach involves partitioning a trained  \texttt{ANN} model using ReLU conditions to obtain linear maps for each partition cell. The minterm weights of these linear maps are interpreted as bit numbers. From the resulting three-dimensional bit tensor, we apply techniques from Triadic Concept Analysis to extract triadic concepts, which are then interpreted as logic trees.

The main benefit of our approach is the ability to interpret an \texttt{ANN} model using logic. Logic provides a powerful toolset for analyzing how attributes interact with each other.

Our approach relies on using all minterm values as network input. Since there are $2^n$ minterms for $n$ object attributes,   $n$ is restricted to be relatively small. In future work, we will strive to adapt this approach to more complex neural networks.

\bibliographystyle{splncs04}
\bibliography{paper}

\end{document}